\documentclass[10pt,twocolumn,letterpaper]{article}

\usepackage[pagenumbers]{iccv} 

\definecolor{iccvblue}{rgb}{0.21,0.49,0.74}
\usepackage[pagebackref,breaklinks,colorlinks,allcolors=iccvblue]{hyperref}

\definecolor{azure}{RGB}{240, 255, 255}

\usepackage{booktabs}
\usepackage{diagbox}
\usepackage{tabularx}
\usepackage{arydshln}
\usepackage{multirow}
\usepackage{amsfonts}
\usepackage{pifont}
\usepackage{amsmath}
\usepackage{dsfont}
\usepackage{subcaption}
\usepackage{threeparttable}
\usepackage{listings}
\usepackage{xcolor}
\usepackage[accsupp]{axessibility} 
\usepackage{wrapfig}

\lstdefinestyle{mystyle}{
    backgroundcolor=\color{gray!10}, 
    commentstyle=\color{gray},       
    keywordstyle=\color{blue},       
    stringstyle=\color{red},         
    basicstyle=\ttfamily\footnotesize, 
    breakatwhitespace=false,         
    breaklines=true,                 
    captionpos=b,                    
    frame=tb,                        
    keepspaces=true,                 
    numbers=left,                     
    numbersep=5pt,                   
    numberstyle=\tiny\color{gray},   
    showspaces=false,                
    showstringspaces=false,          
    showtabs=false,                  
    tabsize=4                        
}

\newcommand*\numcircledmod[1]{\raisebox{.5pt}{\textcircled{\raisebox{-.9pt} {#1}}}}

\newcommand{\cmark}{\ding{51}}%
\newcommand{\xmark}{\ding{55}}%

\RequirePackage{xspace}
\makeatletter
\DeclareRobustCommand\onedot{\futurelet\@let@token\@onedot}
\def\@onedot{\ifx\@let@token.\else.\null\fi\xspace}

\newcommand{\method}{{ESTimator}\xspace}

\def\paperID{8205} 
\def\confName{ICCV}
\def\confYear{2025}

\title{Online Generic Event Boundary Detection}

\author{
Hyungrok Jung$^{1*}$ \quad Daneul Kim$^{2*\S\ddagger}$ \quad Seunggyun Lim$^{1}$ \quad Jeany Son$^{3\dagger\S}$ \quad Jonghyun Choi$^{2\dagger\ddagger}$ \\ [0.2em]
$^1$GIST \quad\quad $^2$Seoul National University \quad\quad $^3$POSTECH\\
{\tt\small jhrock2001@gm.gist.ac.kr} \quad {\tt\small carpedkm@snu.ac.kr} \quad {\tt\small sk000514@gm.gist.ac.kr} \vspace{-0.5mm}\\
{\tt\small jeany@postech.ac.kr} \quad {\tt\small jonghyunchoi@snu.ac.kr} 
}
\vspace{-10mm}

\begin{document}
\maketitle
\let\thefootnote\relax\footnotetext{\hspace{-2em}${*}$ Equal Contribution, ${\dagger}$ Corresponding Authors}
\let\thefootnote\relax\footnotetext{\hspace{-2em}$\S$ Work done while at GIST.}
\let\thefootnote\relax\footnotetext{\hspace{-2em}$\ddagger$ Daneul Kim is with CSE in SNU, and was previously with IPAI. Jonghyun Choi is with ECE, ASRI, and IPAI in SNU.}

\begin{abstract}
Generic Event Boundary Detection (GEBD) aims to interpret long-form videos through the lens of human perception. 
However, current GEBD methods require processing complete video frames to make predictions, unlike humans processing data online and in real-time.
To bridge this gap, we introduce a new task, Online Generic Event Boundary Detection (On-GEBD), aiming to detect boundaries of generic events immediately in streaming videos. 
This task faces unique challenges of identifying subtle, taxonomy-free event changes in real-time, without the access to future frames.
To tackle these challenges, we propose a novel On-GEBD framework, \method{}, inspired by Event Segmentation Theory (EST)~\cite{est} which explains how humans segment ongoing activity into events by leveraging the discrepancies between predicted and actual information. 
Our framework consists of two key components: the Consistent Event Anticipator (CEA), and the Online Boundary Discriminator (OBD).
Specifically, the CEA generates a prediction of the future frame reflecting current event dynamics based solely on prior frames. 
Then, the OBD measures the prediction error and adaptively adjusts the threshold using statistical tests on past errors to capture diverse, subtle event transitions.
Experimental results demonstrate that \method{} outperforms all baselines adapted from recent online video understanding models and achieves performance comparable to prior offline-GEBD methods on the Kinetics-GEBD and TAPOS datasets. 
\end{abstract}

\section{Introduction}
\label{sec:intro}
\begin{figure}[t!]
    \centering
    \includegraphics[width=1\columnwidth]{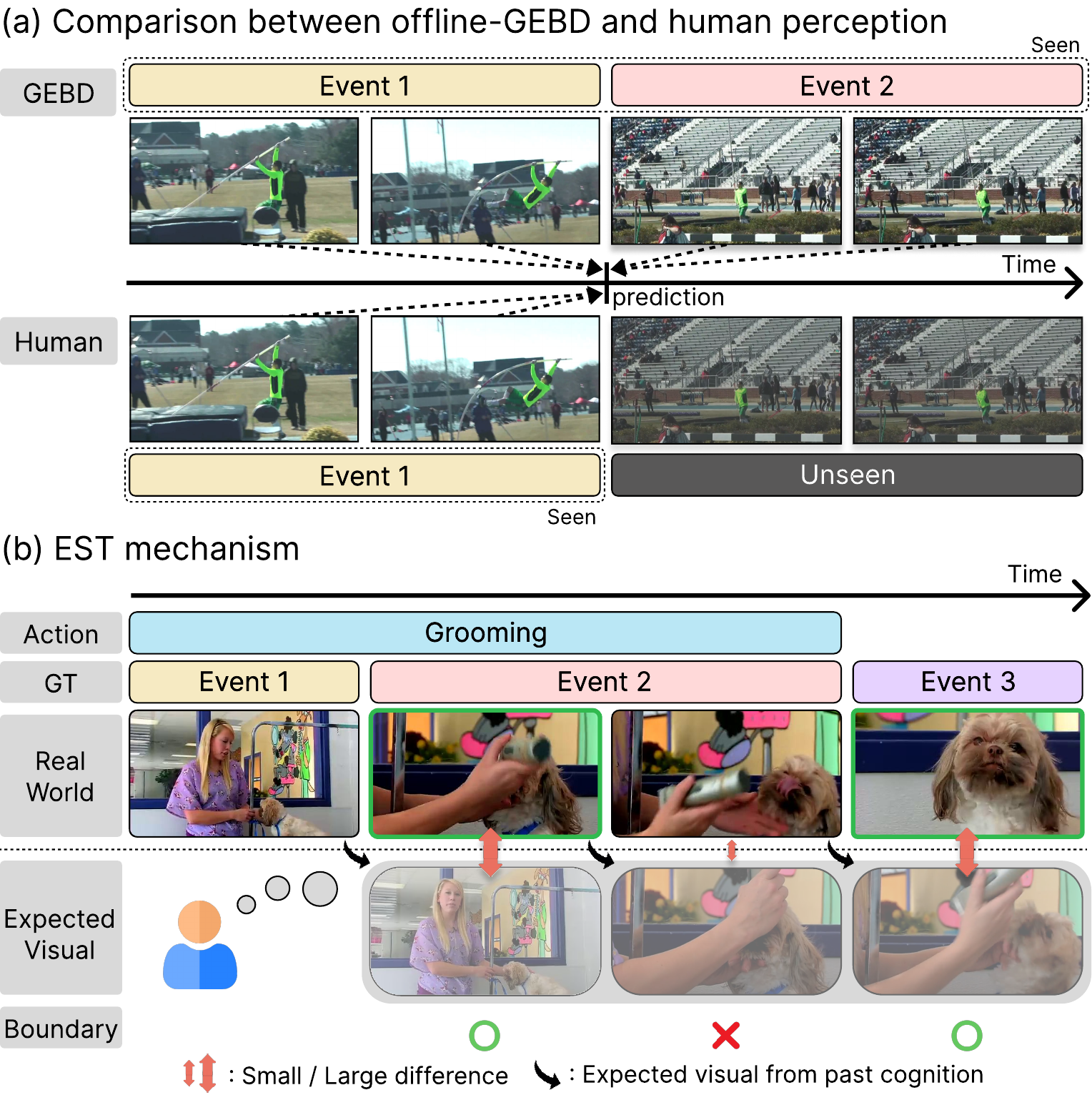}
    \caption{\textbf{Comparison of offline-GEBD and human perception with illustration of Event Segmentation Theory (EST).} (a) In a conventional GEBD task, all the event boundaries are determined by utilizing all past and future frames. However, human segments event sequentially relying only on visuals available at the current moment. (b) The illustration of EST shows how humans perceive events. When we perceive visuals, we naturally expect continuous visuals to be recognized. When a significant difference from the given visual input occurs, we perceive it as an event boundary.
    }
    \vspace{-0.5em}
    \label{fig:motivation}
\end{figure}
In the era of diverse video content from platforms such as YouTube, TikTok and Netflix, the growing importance of processing long-form videos has spurred a growth of interest in video understanding tasks.
While recent research has actively investigated tasks such as action recognition and action detection~\cite{actionformer, bsn++, sola, testra, cag-qil, videoswin, timesformer}, these studies typically focus on a limited set of pre-defined action classes and short, trimmed video clips. 
Consequently, the field of long-form video understanding, which involves analyzing extended video content with complex narratives and diverse actions, remains largely unexplored.

In this regard, Generic Event Boundary Detection (GEBD)~\cite{gebd_original} introduces a new perspective to understand long-form videos.
Originating from cognitive science, the term \textbf{\emph{event}} is an entity of which humans naturally segment continuous visual information, maintaining semantic continuity. 
As humans perceive broad and diverse visuals, events have the characteristic of being taxonomy-free with various levels of granularity.
For example, as illustrated in Figure~\ref{fig:motivation}(b), a pre-defined action like \textit{grooming} can be partitioned into multiple sub-events, while frames that do not correspond to any pre-defined action label still be delineated as generic events.
By aiming to detect changes between these events, GEBD attempts to analyze long-form videos from a human-like perspective. 

However, the current GEBD task aims to identify multiple event boundaries at once, within a fully given chunk of video, which differs significantly from how humans perceive events in an \emph{online} manner (refer to Figure~\ref{fig:motivation}(a)).
By relying on both past and future frames to determine event boundaries, GEBD does not fully reflect natural human perception, as \emph{humans process visual information from time to time without looking at the future}.
To address this limitation and to closely mimic human cognition, we propose a new challenging task, \textbf{\emph{Online Generic Event Boundary Detection (On-GEBD)}}.
In On-GEBD, the model must process a streaming video and decide immediately whether each incoming frame is an event boundary, relying solely on past and present information without access to future frames.

On-GEBD not only inherits the challenges from previous offline-GEBD but also faces more complex challenges due to the limited information (\ie past-only information) when determining event boundaries.
Prior offline-GEBD methods effectively address the challenges in detecting subtle semantic changes of different events by generating a multi-level difference map among several frames~\cite{ddm_net} or creating the temporal similarity matrix for the entire video frame~\cite{uboco}.
However, due to their reliance on a complete sequence of video frames, these methods are not suitable for On-GEBD, where only past information is available to determine whether a streamed frame is a boundary or not.
Also, since an online setting requires immediate determination of boundary as each frame is streamed, it is essential to devise a method specially tailored for On-GEBD.

To address these issues, we propose a simple yet effective framework \emph{\method{}} inspired by Event Segmentation Theory (EST) in cognitive science.
The EST states that humans continuously make predictions consistent with an ongoing event and detect changes when these predictions diverge from the actual information (refer to Figure~\ref{fig:motivation}(b)).
We design a Consistent Event Anticipator (CEA) module to reflect the essence of the EST.
To make CEA predict consistent events robustly, we propose the two training objectives to derive the model to predict visual information consistent with the current ongoing event.
\method{} detects event boundaries by assessing the discrepancy between the visual information predicted by CEA and the actual information.
Moreover, determining boundaries based on a fixed threshold presents challenges in distinguishing events that are not constrained by a particular taxonomy and have varying degrees of granularity.
Therefore, we propose an Online Boundary Discriminator (OBD) module that determines boundaries by comparing the distribution of visual discrepancies from the surrounding frames.
OBD stores historical discrepancies in a fixed-size queue and conducts statistical testing to provide a dynamic threshold that reflects the surrounding context. 


Our method successfully tackles the unique challenges of On-GEBD and shows its effectiveness on two GEBD benchmark datasets, Kinetics-GEBD~\cite{gebd_original} and TAPOS~\cite{tapos}.
We show that our model not only outperforms baseline methods that utilize existing online video understanding methods~\cite{testra, simon, oadtr, miniroad}, but also achieves comparable results to prior offline methods that are tested under the original GEBD setting.

We summarize our contribution as follows:
\begin{itemize}
    \item We present a new challenging task, On-GEBD, designed to align closer to the actual human perception.
    \item To address the unique challenges posed by On-GEBD, we propose a novel framework, \emph{\method{}}, comprising with a Consistent Event Anticipator (CEA) and Online Boundary Discriminator (OBD).
    \item Our model outperforms various baselines based on traditional online video models and achieves performance on par with offline setting~\cite{gebd_original,coseg}.
\end{itemize}

\begin{figure*}[t]
    \centering
    \includegraphics[width=1.0\textwidth]{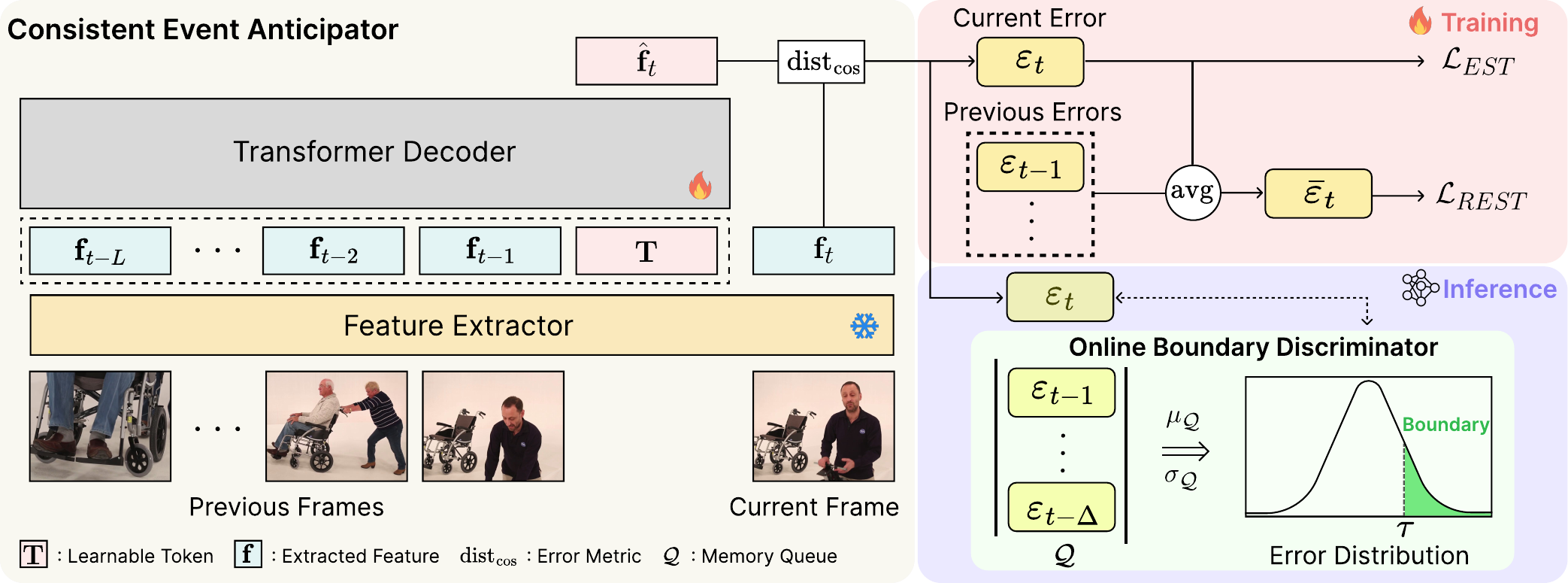} 
        \caption{\textbf{Overview of our \textit{ESTimator} framework.
    } 
    Our framework consists of three major components: 
    Consistent Event Anticipator (CEA) which generates a consistent future frame feature using a learnable token (Left). 
    EST-inspired training objective that accumulates frame-level (EST loss) and region-level (REST loss) prediction errors derived from the discrepancy between the generated future frame from CEA with the actual input frame (Upper right).
    Online Boundary Discriminator (OBD) with a queue that stores past error prediction values to conduct statistical testing on the error derived from the current input frame for inference (Lower right).
    }
    \label{fig:overview}
\end{figure*}

\section{Related Work}
\label{sec:related_work}

\paragraph{Generic Event Boundary Detection.}
Generic Event Boundary Detection (GEBD), introduced in~\cite{gebd_original}, aims to detect event boundaries aligning with human perception. 
Unlike traditional video understanding tasks (\eg, TAL~\cite{re2tal, actionformer, afsd}, Action Recognition~\cite{tran2015learning, huang2022tada}), GEBD deals with continuous semantics without taxonomy, enabling the understanding of complex video.
Recent GEBD works aim to detect boundaries by processing entire videos or analyzing surrounding frame context. 
For instance, UBoCo~\cite{uboco} uses temporal similarity matrices and contrastive learning, while DDM-Net~\cite{ddm_net} employs progressive attention on multi-level dense difference maps.
CoSeg~\cite{coseg}, inspired by cognitive modeling~\cite{est}, introduces boundary detection through event reconstruction.
While offline-GEBD methods~\cite{uboco, ddm_net, e2e, coseg, selfsup, lcvsl, efficientgebd, dybnet,flowgebd,temporal_perceiver,sc-transformer} have shown improvements, our work adapts GEBD to an online setting, addressing the challenge of making instant decisions without future frames.

\vspace{-2mm}
\paragraph{Video Understanding.}
Video understanding encompasses various tasks, including Temporal Action Detection (TAD)~\cite{e2e-tad, tridet, g-tad}, Temporal Action Localization (TAL)~\cite{re2tal, actionformer, afsd}, and Video Instance Segmentation (VIS)~\cite{yang2019video,ke2023mask}.
Recent advancements utilize transformers~\cite{actionformer}, memory structures~\cite{tallformer}, and end-to-end approaches~\cite{afsd}.
However, these tasks often rely on pre-defined classes, limiting real-world applicability. 
While open-set tasks in TAL~\cite{bao2022opental} have been proposed, they still focus on class-based localization. 
GEBD~\cite{gebd_original} addresses this limitation by dealing with taxonomy-free events.

\vspace{-3mm}
\paragraph{Online Video Understanding.}
Online video understanding focuses on streaming content, sharing commonalities with On-GEBD.
Tasks include Online Action Detection (OAD)~\cite{oad}, Online Temporal Action Localization (On-TAL)~\cite{cag-qil}, and Online Detection of Action Start (ODAS)~\cite{odas}.
Recent works like LSTR~\cite{lstr}, TeSTra~\cite{testra}, and MiniROAD~\cite{miniroad} have demonstrated the effectiveness of transformers and GRUs in OAD. 
For On-TAL, methods like CAG-QIL~\cite{cag-qil} and OAT~\cite{sliding-ontal} have been proposed.
While ODAS may seem similar to On-GEBD, it differs significantly as it deals with fixed action classes. 
On-GEBD focuses on detecting subtle changes among taxonomy-free events, requiring a distinct approach from conventional online video understanding methods.

\section{Online Generic Event Boundary Detection}
\label{sec:ongebd}
In this section, we present a new task of Online Generic Event Boundary Detection (On-GEBD), focusing on the identification of event boundaries in streaming videos, which is distinct from the traditional offline-GEBD task.

The offline-GEBD~\cite{gebd_original} considers a video consisting of $N$ frames, \( V = \{v_t\}_{t=1}^{N} \), segmented into multiple events with $M$ distinct generic event boundaries denoted as $\mathcal{B} = \{ b_j\}_{j=1}^{M}$, where $b_j$ represents the timestamp of the $j$-th event boundary.
However, unlike the human perception process, where incoming visual information is segmented into events instantaneously, the offline-GEBD task allows to utilize all frames in $V$ to determine boundaries \(b_j\).

The On-GEBD task imposes two key constraints on the conventional offline-GEBD:
(1) the video is streamed sequentially frame-by-frame;
(2) the model must make immediate decisions for each incoming frame on whether $v_t$ is an event boundary as soon as the frame is received.
These constraints limit the model to use only the past and current frames (\ie, $v_1 \sim v_{t}$) for decision-making, without access to the future context of $v_{t+1} \sim v_N$.
This online, constrained setting makes the model to closely resemble the real-time, causal nature of human event perception.

By virtue of its online setting, On-GEBD presents a novel and more demanding set of challenges than the previous offline-GEBD task.
The absence of future context exacerbates the issues inherited from offline-GEBD in detecting diverse, generic event boundaries.
Since the model is compelled to make instant decisions based on limited information, it must balance rapid boundary detection with the risk of false positives.
Consequently, On-GEBD necessitates the development of novel algorithms that robustly detect the event transitions in streaming visual data with low-latency, which closely aligns with human perception.

\section{Method}
\label{sec:method}

Our framework, \emph{\method{}}, addresses the unique challenges of On-GEBD by drawing inspiration from the Event Segmentation Theory (EST), which explains human event perception~\cite{event_perception}.
\method{} incorporates two key principles from the EST: (1) humans continuously anticipate consistent information about ongoing events; (2) perceive event changes when there is a significant discrepancy between anticipated and actual information (refer to Figure~\ref{fig:motivation}(b)).
These principles offer an effective approach to overcome the key challenges of On-GEBD.
The continuous anticipation of event information allows the model to make decisions, even without access to future frames.
Moreover, by focusing on the trends in discrepancies between anticipated and actual information, the model can effectively detect subtle and diverse event boundaries without relying on pre-defined taxonomies, thereby addressing the ambiguous and varied nature of generic events in streaming video.

Regarding the principles, our framework consists of two components: Consistent Event Anticipator (CEA) and the Online Boundary Discriminator (OBD).
CEA is trained with two novel training objectives to anticipate consistent visual information of the ongoing event robustly.
Since the consistent information anticipated from CEA diverges from actual information at event boundaries, we leverage this discrepancy as a reliable cue for event boundary detection in online scenario.
For clarity, we define the term \textbf{\emph{error}} to denote the discrepancy between the actual and anticipated visual information in the following section.
Also, OBD incorporates prior errors to effectively detect diverse forms of event boundaries, enabling more refined and precise detection based on prior semantic changes.
In the following section, we explore each component in detail.

\subsection{Consistent Event Anticipator}
\label{sec:anticipator}
Transformers have proven their effectiveness in handling sequential data for video understanding tasks, particularly in predicting future states based on past and present information~\cite{Swinlstm, seqformer, futuretrans, anticipativetrans, bert4rec}.
Especially, transformer decoders endowed with auto-regressive causal masking excel at next-information prediction, as exemplified by Large Language Models (LLMs).
In this context, we utilize the transformer decoder~\cite{transformer} in Consistent Event Anticipator (CEA) to construct a model that anticipates consistent frame information aligned with ongoing events.

\vspace{-2mm}\paragraph{Transformer Decoder.}Following a previous GEBD work~\cite{uboco}, we first extract frame features, ${\mathbf{f}_t \in {\mathbb{R}}^D}$, using a pre-trained ResNet-50 image encoder~\cite{resnet}.
We concatenate a single learnable token $\mathbf{T}\in {\mathbb{R}}^{D}$ with $L$ extracted frame features $\mathbf{F}_{t} = \{{\mathbf{f}_i}\}_{i=t-L}^{t-1}$ and forward them into transformer decoder layers $\mathcal{M}_\theta$ to predict streamed frame features.

\begin{equation}
\label{eq:transformer_decoder}
    \begin{aligned}
    &\mathbf{X}_t = \text{concat}( \mathbf{F}_{t}, \hspace{2pt} \mathbf{T}),  \\    &\hat{\mathbf{f}}_t = \mathcal{M}_\theta (\mathbf{X}_t),       
    \end{aligned}
\end{equation}
In Eq.~\ref{eq:transformer_decoder}, $\hat{\mathbf{f}}_t$ denotes the output of the learnable token $\mathbf{T}$ after processing through $\mathcal{M}_\theta$, which encapsulates a prediction for the upcoming frame feature (refer Figure~\ref{fig:overview}).
With the causal-attention mask in the transformer decoder, we ensure that succeeding tokens only attend to the preceding ones. 

\vspace{-2mm}\paragraph{Objective Functions.}The main objective of CEA is to maximize errors at semantically inconsistent event boundaries while minimizing errors within consistent event segments, thereby embodying the core principles of EST.
We preliminarily define the error $\varepsilon_t$ between the prediction $ \hat{\mathbf{f}}_t$ and the actual frame feature $\mathbf{f}_t$ with the cosine distance, and scale it between 0 and 1 as follows:
\vspace{-0.1cm}
\begin{equation}
\label{eq:error}
    \varepsilon_t = \text{dist}_{\text{cos}}(\mathbf{f}_t, \hat{\mathbf{f}}_t) = \frac{1}{2} \left(1 - \frac{{\mathbf{f}_t \cdot \hat{\mathbf{f}}_t}}{\|\mathbf{f}_t\| \|\hat{\mathbf{f}}_t\|} \right).
\end{equation}
To achieve the objective of CEA following EST, we use a binary cross-entropy loss for our \textbf{EST loss}, as follows:
\vspace{-0.1cm}
\begin{equation}
\label{eq:est_loss}
    \begin{aligned}
    \mathcal{L}_{EST} (\varepsilon_t, t) = -y_t\log\varepsilon_{t} - (1 - y_t)\log(1 - \varepsilon_{t}),
    \end{aligned}
\end{equation}
where $y_t=1$ if $t \in \mathcal{B}$, otherwise 0.
The EST loss encourages $\mathcal{M}_\theta$ to maximize the errors at event boundaries while minimizing them elsewhere, effectively distinguishing boundary frames from non-boundary frames.

However, strict frame-wise binary supervision would be sub-optimal in videos where consecutive frames contain continuous semantic flow.
This is because consecutive frames are smoothly connected without abrupt changes, except shot changes.
To address this issue, we propose a region-level training scheme that considers temporal context flow.
Since this training scheme shares the mechanism of the EST loss, we named it \textbf{REST loss} (\textbf{R}egion \textbf{EST loss}).
Incorporating errors derived from nearby regions, REST loss aims to give soft supervision of the abrupt label transitions in the near future.
Assuming the size of a region as $K$, we collect the series of errors $\varepsilon_{t-K}, \dots, \varepsilon_t$ from the inputs $\mathbf{X}_{t-K}$, $\dots$, $\mathbf{X}_{t}$, and compute the average of these consecutive past errors $\Bar{\varepsilon}_t$ as follows:
\vspace{-0.2cm}
\begin{equation}
            \Bar{\varepsilon}_t = \frac{1}{K} \sum_{i=t-K}^t \varepsilon_i.
\end{equation}
The REST loss is then defined as follows:
\vspace{-0.1cm}
\begin{equation}
            \mathcal{L}_{REST} (\varepsilon_t,t) =  \mathcal{L}_{EST}(\Bar{\varepsilon}_t, t),
            \label{eq:rest}
\end{equation}
where the boundary label at $t$ is used for the loss.
The REST loss allows the predicted frame features, ${\hat{\mathbf{f}}}_{t-K} \dots {\hat{\mathbf{f}}}_{t}$, to be softly trained with additional future information, making them less sensitive to noise and more effective in anticipating future events.

Our final loss function is computed as a weighted sum of two losses, EST and REST losses, as follows:
\vspace{-0.2cm}
\begin{equation}
    \mathcal{L} (\varepsilon_t, t) =   \alpha \cdot \mathcal{L}_{REST}(\varepsilon_t, t) + \sum_{i=t-K}^t \mathcal{L}_{EST}(\varepsilon_i, i).
    \vspace{-0.1cm}
\label{eq:final_loss}
\end{equation}
In our experiments, we set the hyperparameter $\alpha$ to 0.5.

\vspace{-2mm}\paragraph{Batch-wise Loss Weighting.}
In the video, frames that correspond to boundaries account for a significantly smaller proportion compared to non-boundary frames.
Previous offline-GEBD studies utilize balanced samplers~\cite{gebd_original, ddm_net} or weighted loss terms~\cite{maskedgebd} with manually tuned values to alleviate issues arising from such imbalanced data distribution.
To achieve a similar effect during training, we utilize batch-wise loss weighting technique.
We first calculate the ratio of boundary and non-boundary targets within a single batch.
This ratio is then multiplied by the loss calculated for boundary targets, allowing the batch-wise loss weighting to balance the impact of boundary and non-boundary samples during training. 
This approach not only eliminates the need for manually tuned scaling values but also aims to achieve a similar effect to that of using a batch sampler. 

\subsection{Online Boundary Discriminator}
\label{sec:obd}
Conventional offline-GEBD methods typically rely on either static thresholds~\cite{ddm_net, gebd_original, maskedgebd} or dynamic criteria based on peak detection~\cite{coseg} to identify boundaries by analyzing the entire sequence of frames—including future ones.
However, these approaches are not well-suited to the On-GEBD: static thresholds fail to capture diverse form of semantic changes, and peak detection is impractical due to its reliance on future frame information.
These limitations highlight the need for a novel approach tailored to the immediacy and dynamic nature of On-GEBD.

\begin{wrapfigure}{r}{0.41\columnwidth}
    \vspace{-0.3cm}
    \hspace{-0.5cm}
    \includegraphics[width=0.43\columnwidth, trim=2 0 2 0, clip]{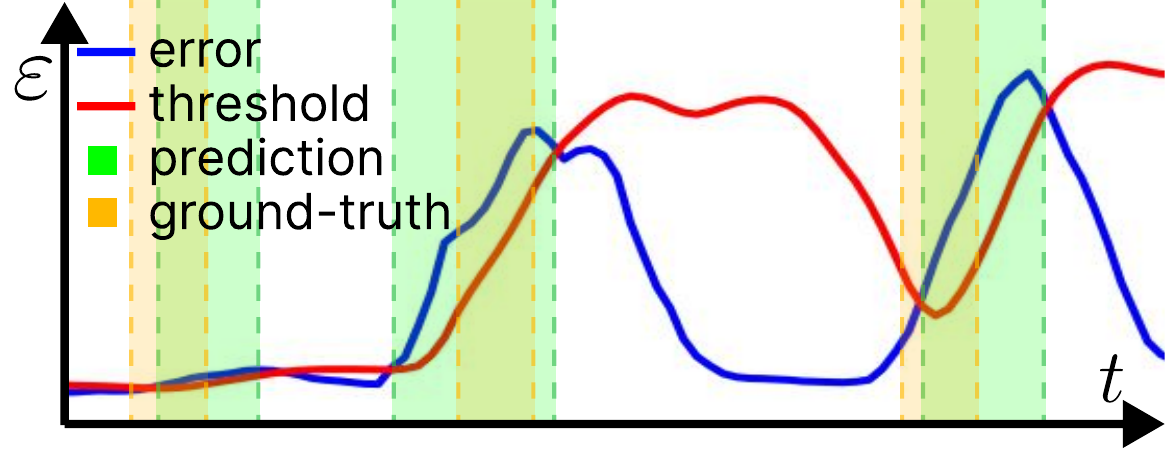}
    \vspace{-0.3cm}
    \captionsetup{
          margin={-0.4cm,0cm}   
      }
    \caption{An illustration of how Online Boundary Detector (OBD) applies a dynamic threshold to capture diverse event transitions.}
    \label{fig:obd_explain}
    \vspace{-0.4cm}
\end{wrapfigure}
To address these unique challenges, we propose an Online Boundary Discriminator (OBD), which dynamically discriminates event boundaries by leveraging recent semantic flow.
First, OBD stores errors up to the current frame in a fixed-size memory queue, denoted as $\mathcal{Q}$, which is managed in a First-In, First-Out (FIFO) manner.
When the $t$-th frame $v_t$ is streamed, $\mathcal{Q}$ contains past errors \(\varepsilon_{t-\Delta}\), $\dots$, \(\varepsilon_{t-1}\), where $\Delta$ is the size of the $\mathcal{Q}$.
Using the errors stored in a memory queue, OBD conducts statistical testing with the Gaussian distribution derived from errors in the $\mathcal{Q}$.
By normalizing the incoming error $\varepsilon_t$ with statistical information derived from $\mathcal{Q}$, the OBD determines a boundary if the incoming error is identified as an outlier compared to the close region errors.
Figure~\ref{fig:obd_explain} exemplifies how the underlying mechanism of OBD dynamically modulates the detection threshold.
\begin{equation}
    \begin{aligned}
        &\zeta_t=\frac{\varepsilon_t-\mu_{\mathcal{Q}}}{\sigma_{\mathcal{Q}}},\\
        &\text{OBD}(\mathcal{Q},  \hspace{0.3em} \varepsilon_t)= \mathds{1} \hspace{0.1em} [\zeta_t > \tau ],
    \end{aligned}
    \label{eq:obd_equation}
\end{equation}
In Eq.~\ref{eq:obd_equation}, $\tau$ is a threshold value for an indicator function $\mathds{1}[\cdot]$ and $\zeta_t$ is the normalized error at frame $t$ by the mean $\mu_{\mathcal{Q}}$ and the standard deviation $\sigma_{\mathcal{Q}}$ derived from errors stored in $\mathcal{Q}$ (refer lower right side of the Figure~\ref{fig:overview}). 

In our experiments, we set $\tau$ to 1.5 empirically.
By dynamically adjusting the error threshold through OBD, our ESTimator successfully identifies a wide range of event boundaries, as demonstrated in Figure~\ref{fig:quali_1}.

\begin{table*}[t]
    \centering
     \caption{\textbf{Quantitative comparison with other online baselines in On-GEBD task.} 
    BC denotes a binary classifier attached to the last layer of each model to solve On-GEBD. We denoted with \textbf{bold} for the highest F1 score and second with \underline{underline}.
    }
    \resizebox{\textwidth}{!}{
    \begin{tabular}{cccccccccccccc}
    \hline
    \toprule
     Dataset & Rel. Dis. threshold & 0.05 & 0.1 & 0.15 & 0.2 & 0.25 & 0.3 & 0.35 & 0.4 & 0.45 & 0.5 & Avg \\
    \midrule 
       \multirow{5}{*}{Kinetics-GEBD}& TeSTra - BC~\cite{testra} & 0.438 & 0.488 & 0.521 & 0.545 & 0.564 & 0.580 & 0.593 & 0.604 & 0.614 & 0.622 & 0.557 \\
   & Sim-On - BC~\cite{simon}& 0.461 & 0.534 & 0.579 & 0.610 & 0.633 & 0.651 & 0.664 & 0.675 & 0.685 & 0.692 & 0.618 \\
   & OadTR - BC~\cite{oadtr} & 0.474 & 0.512 & 0.535 & 0.552 & 0.565 & 0.575 & 0.583 & 0.590 & 0.596 & 0.601 & 0.558 \\
   & MiniROAD - BC~\cite{miniroad} & \underline{0.569} & \underline{0.622} & \underline{0.649} & \underline{0.675} & \underline{0.691} & \underline{0.704} & \underline{0.714} & \underline{0.722} & \underline{0.729} & \underline{0.735} & \underline{0.681} \\
   \cmidrule{2-14}
   & \textbf{\method{} (Ours)} & \bf 0.620 & \bf 0.687 & \bf 0.724 & \bf 0.746 & \bf 0.762 & \bf 0.774 & \bf 0.782 & \bf 0.789 & \bf 0.795 & \bf 0.799 & \bf 0.748 \\
    \midrule
    \midrule
    \multirow{5}{*}{TAPOS}& TeSTra - BC~\cite{testra} & 0.364 & 0.417 & 0.452 & 0.478 & 0.496 & 0.511 & 0.523 & 0.533 & 0.542 & 0.550 & 0.487 \\
    & Sim-On - BC~\cite{simon} & 0.225 & 0.269 & 0.303 & 0.329 & 0.350 & 0.367 & 0.381 & 0.394 & 0.405 & 0.415 & 0.344 \\
    & Oad-TR - BC~\cite{oadtr} & 0.263 & 0.319 & 0.361 & 0.394 & 0.422 & 0.445 & 0.465 & 0.483 & 0.497 & 0.510 & 0.416 \\
    & MiniROAD - BC~\cite{miniroad} & \bf 0.422 & \bf 0.472 & \bf 0.502 & \underline{0.522}  & \underline{0.537}  & \underline{0.549}  & \underline{0.558}  & \underline{0.566}  & \underline{0.572}  & \underline{0.578} & \underline{0.528} \\
    \cmidrule{2-14}
    & \textbf{\method{} (Ours)} & \underline{0.394} & \underline{0.455} & \underline{0.499} & \bf 0.532 & \bf 0.558 & \bf 0.578 & \bf 0.594 & \bf 0.608 & \bf 0.619 & \bf 0.629 & \bf 0.547 \\
    \bottomrule
    \hline
    \end{tabular}
    }
        \label{tab:online_only}
\end{table*}

\begin{table*}[t]
    \centering
    \caption{\textbf{Quantitative comparison with offline methods.} Note that we report the performance of the models in offline setting from their original literature.
    Also, we indicate the highest F1 score with \textbf{bold}, second with \underline{underline} and third with \textdagger. 
    }
    \resizebox{\textwidth}{!}{
    \begin{tabular}{cccccccccccccccc}
    \hline
    \toprule
    Dataset&Setting&Supervision&Rel. Dis. threshold & 0.05 & 0.1 & 0.15 & 0.2 & 0.25 & 0.3 & 0.35 & 0.4 & 0.45 & 0.5 & Avg \\
    \midrule
    \multirow{11}{*}{\rotatebox[origin=c]{90}{Kinetics-GEBD}}&\multirow{10}{*}{Offline}&\multirow{5}{*}{Supervised}& BMN~\cite{bmn} & 0.186 & 0.204 & 0.213 & 0.220 & 0.226 & 0.230 & 0.233 & 0.237 & 0.239 & 0.241& 0.223\\
    &&& BMN-StartEnd~\cite{gebd_original}&0.491 & 0.589 & 0.627 & 0.648 & 0.660 & 0.668 & 0.674 & 0.678 & 0.681 & 0.683& 0.640\\
    &&& TCN-TAPOS~\cite{gebd_original}& 0.464 & 0.560 & 0.602 & 0.628 & 0.645 & 0.659 & 0.669 & 0.676 & 0.682 & 0.687 & 0.627\\
    &&& TCN~\cite{TCN}&0.588 & 0.657 & 0.679 & 0.691 & 0.698 & 0.703 & 0.706 & 0.708 & 0.710 & 0.712 & 0.685\\
    &&& PC~\cite{gebd_original} & 0.625 & \bf 0.758 & \bf 0.804 & \bf 0.829 & \bf 0.844 & \bf 0.853 & \bf 0.859 & \bf 0.864 & \bf 0.867 & \bf 0.870 & \bf 0.817\\
    \cmidrule{3-15}
    &&\multirow{4}{*}{Unsupervised} & SceneDetect~\cite{gebd_original} &0.275 & 0.300 & 0.312 & 0.319 & 0.324 & 0.327 & 0.330 & 0.332 & 0.334 & 0.335 & 0.318 \\
    &&& PA-Random~\cite{gebd_original} &0.336 & 0.435 & 0.484 & 0.512 & 0.529 & 0.541 & 0.548 & 0.554 & 0.558 & 0.561& 0.506 \\
    &&& PA~\cite{gebd_original} &0.396 & 0.488 & 0.520 & 0.534 & 0.544 & 0.550 & 0.555 & 0.558 & 0.561 & 0.564 & 0.527\\
    &&& CoSeg~\cite{coseg} & \bf 0.656 & \bf 0.758 & \underline{0.783} & \underline{0.794} & \underline{0.799} & \underline{0.803} & \underline{0.804} & \underline{0.806} & \underline{0.807} & \underline{0.809} & \underline{0.782} \\
\cmidrule{2-15}
   &Online&Supervised& \textbf{\method{} (Ours)} & \ \  0.620\textsuperscript{\textdagger} & \  0.687\textsuperscript{\textdagger} & \ 0.724\textsuperscript{\textdagger} &  \ 0.746\textsuperscript{\textdagger} &  \ 0.762\textsuperscript{\textdagger} &  \ 0.774\textsuperscript{\textdagger} &  \ 0.782\textsuperscript{\textdagger} &  \ 0.789\textsuperscript{\textdagger} &  \ 0.795\textsuperscript{\textdagger} &  \ 0.799\textsuperscript{\textdagger} &  \ 0.748\textsuperscript{\textdagger} \\
\midrule
\midrule
       \multirow{9.5}{*}{\rotatebox[origin=c]{90}{TAPOS}}&\multirow{8}{*}{Offline}&\multirow{5}{*}{Supervised}&ISBA~\cite{gebd_original} & 0.106 & 0.170 & 0.227 & 0.265 & 0.298 & 0.326 & 0.348 & 0.348 & 0.348 & 0.348 & 0.330\\
    &&& TCN~\cite{gebd_original} & 0.237 & 0.312 & 0.331 & 0.339 & 0.342 & 0.344 & 0.347 & 0.348 & 0.348 & 0.348& 0.330\\
    &&& CTM~\cite{gebd_original}& 0.244 & 0.312 & 0.336 & 0.351 & 0.361 & 0.369 & 0.374 & 0.381 & 0.383 & 0.385 & 0.350\\
    &&& TransParser~\cite{TCN} & 0.289 & 0.381 & 0.435 & 0.475 & 0.500 & 0.514 & 0.527 & 0.534 & 0.540 & 0.545 & 0.474\\
    &&& PC~\cite{gebd_original} & \bf 0.522 & \bf 0.595 & \bf 0.628 & \bf 0.646 & \bf 0.659 & \bf 0.665 & \bf 0.671 & \bf 0.676 & \bf 0.679 & \bf 0.683 & \bf 0.642\\
    \cmidrule{3-15}
    &&\multirow{3}{*}{Unsupervised} & SceneDetect~\cite{gebd_original} &0.035 & 0.045 & 0.047 & 0.051 & 0.053 & 0.054 & 0.055 & 0.056 & 0.057 & 0.058 & 0.051 \\
    &&& PA-Random~\cite{gebd_original} &0.158 & 0.233 & 0.273 & 0.310 & 0.331 & 0.347 & 0.357 & 0.369 & 0.376 & 0.384 & 0.314 \\
    &&& PA~\cite{gebd_original} & \underline{0.360} & \underline{0.459} & \underline{0.507} & \underline{0.543} & \underline{0.567} & \underline{0.579} & \ 0.592\textsuperscript{\textdagger}  & \ 0.601\textsuperscript{\textdagger}  & \ 0.609\textsuperscript{\textdagger}  & \ 0.615\textsuperscript{\textdagger}  & \ 0.543\textsuperscript{\textdagger} \\
    \cmidrule{2-15}
    &Online&Supervised& \textbf{\method{} (Ours)} & \ 0.394\textsuperscript{\textdagger}  & \ 0.455\textsuperscript{\textdagger}  & \ 0.499\textsuperscript{\textdagger}  & \ 0.532\textsuperscript{\textdagger}  & \ 0.558\textsuperscript{\textdagger}  & \ 0.578\textsuperscript{\textdagger} &  \underline{0.594} & \underline{0.608} & \underline{0.619} & \underline{0.629} & \underline{0.547} \\
    \bottomrule
    \hline
    \end{tabular}
    }
    \label{tab:offline_online_comp}
\end{table*}

\section{Experiment}
\label{sec:exp}

\subsection{Setup}
\label{sec:exp_setup}

\paragraph{Benchmark Dataset.}The Kinetics-GEBD dataset~\cite{gebd_original} is composed of approximately 60K videos selected from the Kinetics-400 dataset~\cite{k400original}.
The selected videos are divided into units of taxonomy-free events by specially trained annotators.
On average, each video contains about 5 different events.
The train, validation and test videos are almost equally distributed with 18,794, 18,813 and 17,725 videos, respectively.
Since annotations for the test set are not available, we report the results evaluated on the validation set, following prior works~\cite{gebd_original, uboco, ddm_net}.
For cross-validation, we randomly partitioned the dataset into training (80\%) and validation (20\%) subsets for the experiments.
TAPOS dataset~\cite{tapos} is composed of Olympic sports videos that are annotated with 21 different actions with  13,094 training action instances and 1,790 validation action instances.
Following~\cite{gebd_original}, we re-purpose TAPOS for the GEBD task by obscuring the action labels of sub-actions.
Additionally, we present results for INRIA~\cite{inria} in supplementary material.
\vspace{-3mm}
\paragraph{Evaluation Metric.}
Relative Distance (Rel.Dis) is a metric that measures the relative difference between the detected boundary timestamps and the ground truth boundary timestamps in GEBD. 
We calculate the metric by dividing the distance between predictions and ground-truths by the union of predicted and ground-truth event instances.
When the model predicts consecutive frames as boundaries, as discussed in~\cite{gebd_original}, we set the center of these frames as the predicted boundary timestamps.
Relative distance (Rel.Dis) of detected timestamps is evaluated under 10 thresholds with intervals of 0.05, ranging from 0.05 to 0.5.
We consider the result to be positive if it is below each threshold. 
Following the prior works, we report the F1 scores on each threshold, as well as their average in our main experiments. 

\vspace{-3mm}
\paragraph{Baseline.}Directly extending the offline-GEBD methods into an online setting poses challenges as they are not designed to process streaming videos.
Therefore, we utilize state-of-the-art online video understanding models, which are natively designed for action detection or localization, as our baseline for On-GEBD.
Specifically, we utilize TeSTra~\cite{testra}, OadTR~\cite{oadtr}, and MiniROAD~\cite{miniroad} from Online Action Detection, and Sim-On~\cite{simon} from Online Temporal Action Localization as our na\"ive baselines.
Only the head of each model is modified to perform binary classification (BC), as it is necessary for the model to distinguish whether a streamed input is a boundary or not.

\vspace{-3mm}
\paragraph{Implementation Details.}We preprocess the video data by sampling at 24~FPS for the Kinetics-GEBD dataset and 6~FPS for the TAPOS dataset.
Following~\cite{uboco}, we use the features extracted from the ResNet-50 encoder, which is pre-trained on ImageNet~\cite{imagenet} with feature dimension $D$ = 2,048.
For our experiments, we employ 3 transformer decoder layers in both datasets.
Additionally, we train our model with a batch size of 512 using the AdamW~\cite{adamw} optimizer in training with a learning rate of 1e-4.

\subsection{Main Result}
\label{sec:exp_main}
Table~\ref{tab:online_only} shows that our framework outperforms baseline models for On-GEBD on both Kinetics-GEBD and TAPOS datasets. 
Traditional online methods like TeSTra-BC, Oad-TR-BC, and Sim-On-BC show a lack of model capacity in learning to discriminate generic event boundaries, limited by their model design to learn pre-defined action classes.
While MiniROAD-BC demonstrates higher performance than other baselines, our approach still surpasses them on both datasets.
The results highlight that simply adapting existing approaches is insufficient; a dedicated method like \method{} is necessary to detect subtle semantic changes during the event transition.

In Table~\ref{tab:offline_online_comp}, we compare our framework with models evaluated in an offline setting. 
\method{} performs on par with or exceeds most offline methods, achieving higher Avg. F1 scores on Kinetics-GEBD—except for PC~\cite{gebd_original} and CoSeg~\cite{coseg}. 
Similarly, on the TAPOS, \method{} outperforms all baselines from the original GEBD, with the sole exception of the PC method~\cite{gebd_original}.

\begin{table}[t!]
    \centering
    \caption{\textbf{Benefit of the proposed components}. As a baseline, we utilize the transformer decoder with binary classifier. We gradually add each of the proposed components to investigate its effect on improving performance. }
    \footnotesize
    \resizebox{1.0 \columnwidth}{!}{
    \begin{tabular}{lccccc}
    \toprule
     \small\multirow{1}{*}{   Method}          & \multirow{1}{*}{\numcircledmod{1} EST} & \multirow{1}{*}{\numcircledmod{2} REST} & \multirow{1}{*}{\numcircledmod{3} OBD} & \small F1 @ 0.05 & \multicolumn{1}{c}{Avg F1} \\ 
      \cmidrule(lr){1-1} \cmidrule(lr){2-4} \cmidrule(lr){5-5}\cmidrule(lr){6-6} 
     ~ Baseline & {\color{red} \xmark} & {\color{red} \xmark}  & {\color{red} \xmark} & {0.483} & {0.607} \\
    ~\numcircledmod{1} &  {\color{ForestGreen} \cmark}&  {\color{red} \xmark}& {\color{red} \xmark} & {0.571} & {0.698}\\
    ~\numcircledmod{2} &  {\color{red} \xmark} & {\color{ForestGreen} \cmark} & {\color{red} \xmark} & {0.504} & {0.654}\\
    \hdashline
    ~\numcircledmod{1}+\numcircledmod{2} & {\color{ForestGreen} \cmark}& {\color{ForestGreen} \cmark}& {\color{red} \xmark} & {0.544} & {0.691}  \\
    ~\numcircledmod{1}+\numcircledmod{3} & {\color{ForestGreen} \cmark}& {\color{red} \xmark}& {\color{ForestGreen} \cmark} & {0.604} & {0.659}  \\
    ~\numcircledmod{2}+\numcircledmod{3} & {\color{red} \xmark}& {\color{ForestGreen} \cmark}& {\color{ForestGreen} \cmark} & {\bf 0.621} & {0.692}  \\
    \hdashline
~\numcircledmod{1}+\numcircledmod{2}+\numcircledmod{3} (Ours) &{\color{ForestGreen} \cmark} & {\color{ForestGreen} \cmark} &{\color{ForestGreen} \cmark} & {0.620} &  {\bf 0.748} \\ \bottomrule
    \\
    \end{tabular}
    }
    \vspace{-1em}
\label{tab:ablation}
\end{table}

\subsection{Ablation Study}
\label{sec:exp_ablation}

\paragraph{Impact of Proposed Components.}
To observe the benefits of each proposed component, we build our components on top of the simple transformer model with a binary classifier, which we refer to as \textit{Baseline} in Table~\ref{tab:ablation}.
The experimental results present the effectiveness of error-based detection of event boundaries, as the model trained with either EST or REST loss consistently outperforms the baseline.

Na\"ively applying the REST loss on top of the EST loss degrades the models' performance, as using both objectives together tends to reduce errors on boundary frames.
Furthermore, the OBD proved to be less effective when applied to a CEA trained exclusively with either EST or REST loss.
The performance gap between our full proposed framework and its ablated versions demonstrates the synergistic effect of our individual components.

\vspace{-2mm}
\paragraph{Metric for Error Calculation.}

\begin{table}[t!]
    \centering
    \caption{\textbf{Comparison on different metric for calculating the error.} We compare Avg. F1 scores using different distance metrics, with min-max normalization applied to each batch during training. Cosine distance is the best choice due to its bounded nature.
    }   
    \footnotesize
    \begin{tabular}{l c}
        \toprule
        \multirow{1}{*}{Error Calculation Metric} & \multirow{1}{*}{Avg F1} \\
        \cmidrule(lr){1-1} \cmidrule(lr){2-2}
        L1 Distance (min/max normalized)  & 0.733 \\
        L2 Distance (min/max normalized)  & 0.733 \\
        KL Divergence (min/max normalized)  & 0.734 \\
        Cosine Distance (Ours) & \textbf{0.748} \\
        \bottomrule
    \end{tabular}
    \label{tab:loss_ablation}
\end{table}

We investigate the impact of the metric used for error computation on the performance of our framework.
Unlike the cosine distance, which is bounded, other widespread metrics such as the L1/L2 distance and the KL Divergence are generally unbounded.
Therefore, to facilitate the application of our proposed EST and REST losses, we experiment these metrics with min-max normalization per batch while training the CEA.
As shown in Table~\ref{tab:loss_ablation}, the cosine distance demonstrates superior performance compared to the L1/L2 distance and the KL divergence.
Furthermore, we note that these alternative metrics remain relatively viable due to our proposed OBD.
OBD provides reliable criteria even for error values with unbounded ranges, thereby mitigating the challenges of selecting an appropriate threshold.

\vspace{-2mm}
\paragraph{Real Time Performance.}
We report a runtime analysis of our method with other baselines in Table~\ref{tab:fps} and demonstrate the feasibility of our approach for real-time processing.
For all methods, we utilize the ResNet-50 encoder pre-trained on ImageNet~\cite{imagenet}, which operates at 181 FPS in our experimental setting.
ESTimator not only demonstrates superior performance, but also achieves higher FPS compared to other transformer-based baselines (\ie, TeSTra-BC, Sim-On-BC, OadTR-BC).
Even with superior performance, its overall FPS is on par with that of MiniROAD-BC, which is based on GRU~\cite{gru} architecture.
We further demonstrate the analysis on computation cost in supplementary material.

\begin{table}[t!]
    \centering
    \caption{\textbf{Comparison of real-time performance (in FPS) with other baselines, which utilize other online video understanding models.}  We denoted with bold for the highest FPS and performance, and underlined for the second highest. \method{} shows the highest Avg. F1 with compatible overall FPS compared to MiniROAD-BC~\cite{miniroad}. All experiments were conducted on a single NVIDIA RTX A6000 GPU.}
    \footnotesize
    \resizebox{1.0 \columnwidth}{!}{
    \begin{tabular}{lcccc}
    \toprule
     \multirow{1}{*}{Method~~~} & \multirow{1}{*}{~~RGB Feat~~} & \multirow{1}{*}{~~Model~~} & ~~Overall~~ & \multicolumn{1}{c}{~~Avg F1~~} \\ 
      \cmidrule(lr){1-1} \cmidrule(lr){2-3} \cmidrule(lr){4-4}\cmidrule(lr){5-5} 
     TeSTra - BC & \multirow{4}{*}{181} & 177 &  72.5 & 0.557\\
     Sim-On - BC &  &  275 &  76.3 & 0.618 \\
     OadTR - BC &   &  100 &  48.9 & 0.558\\
     MiniROAD - BC & & \textbf{3069} & \textbf{99.8} & \underline{0.681} \\
     \textbf{\method{} (Ours)} & &  \underline{481} & \underline{96.3} & \textbf{0.748} \\
\bottomrule
    \\
    \end{tabular}
    }
\vspace{-1.5em}
\label{tab:fps}
\end{table}

\begin{figure*}[t!]
    \centering
    \includegraphics[width=1.0\textwidth]{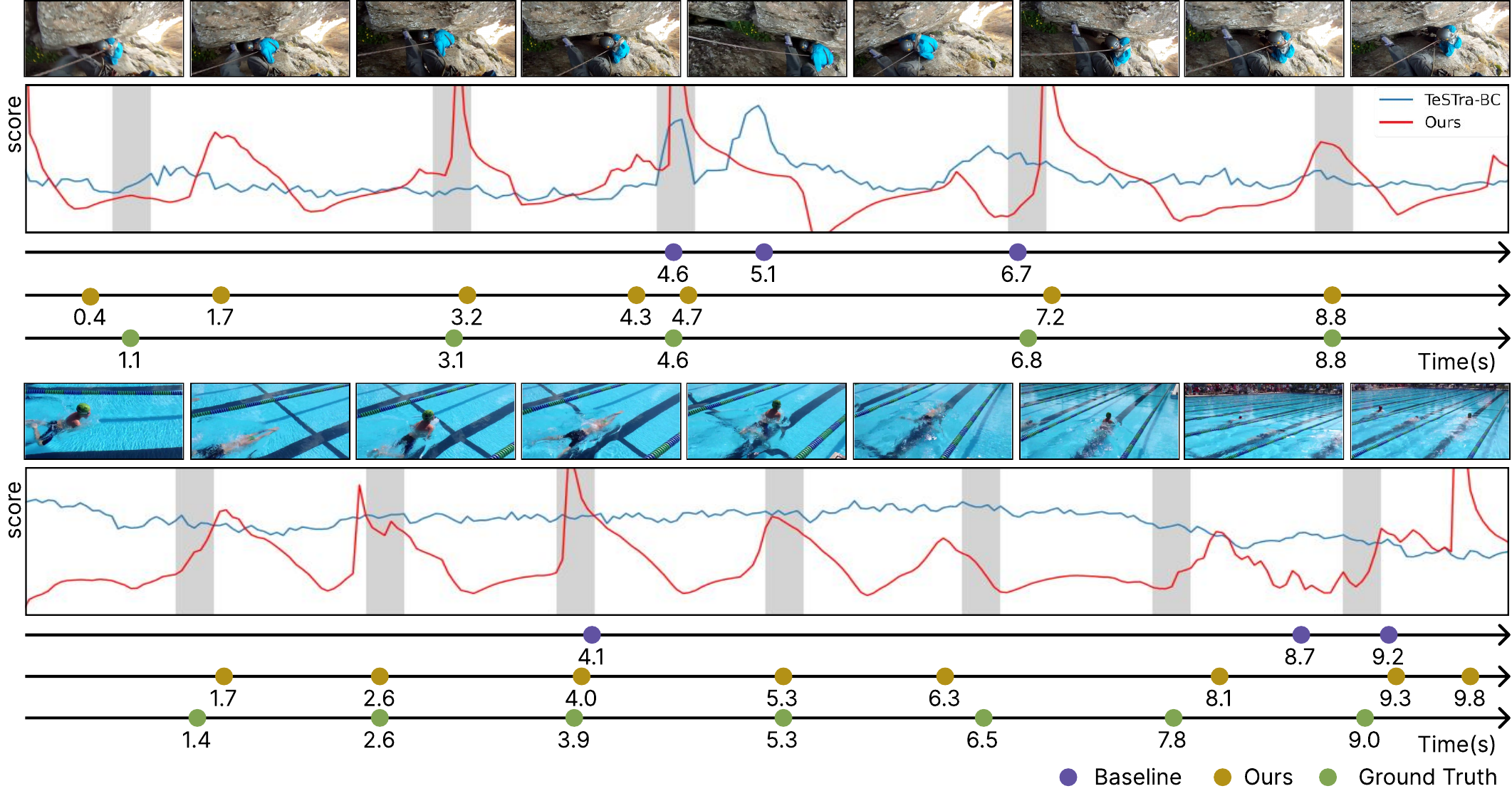} 
        \vspace{-0.4cm}
        \caption{\textbf{Qualitative result.} Comparison between our proposed framework and baseline from other online video understanding task. Note that baseline here refers to TeSTra~\cite{testra} with binary classifier head attached for event boundary detection.
    }
    \vspace{-0.5cm}
    \label{fig:quali_1}
\end{figure*}

\begin{table}[t]
\centering
\caption{\textbf{Video feature comparison.} We compare the performance of our method and baselines using video features extracted from TSN networks pre-trained on Something Something v2 (SS-v2) and Kinetics datasets.}
\footnotesize
 \resizebox{1.\columnwidth}{!}{
    \begin{tabular}{lccccc}
    \toprule
    Method  &     Backbone    &   Avg F1    &     Backbone   &   Avg F1\\
    \cmidrule(lr){1-1} \cmidrule(lr){2-3} \cmidrule(lr){4-5} 
    TeSTra - BC     & \multirow{5}{*}{\shortstack{TSN\\(SS-v2)}}   &0.653      & \multirow{5}{*}{\shortstack{TSN \\ (Kinetics)}}  &0.654 \\
    Sim-On - BC     &       & 0.524  &   &   0.501    \\
    OadTR - BC      &     &  0.700 &       & 0.699      \\
    MiniROAD - BC   &       &0.684  &&0.689  \\
    \textbf{\method{} (Ours)}    &      & \textbf{0.741}  && \textbf{0.744}  \\
    \bottomrule
    \end{tabular}
}   
\label{tab:vidfeat}
\end{table}

\begin{table}[t!]
    \centering
    \caption{\textbf{Comparison on outlier handling in OBD.} Removing outliers from the queue significantly reduces performance, demonstrating our OBD functions as a dynamic threshold adaptation.}
    \footnotesize
    \begin{tabular}{l l}
        \toprule
        \multirow{1}{*}{OBD Setting} & \multirow{1}{*}{Avg F1} \\
        \cmidrule(lr){1-1} \cmidrule(lr){2-2}
        Using Only Inliers & 0.663 \textcolor{blue}{(-11.4\%)} \\
        Full OBD (Ours) & \textbf{0.748} \\
        \bottomrule
    \end{tabular}
    \label{tab:obd_outlier_ablation}
    \vspace{-0.1cm}
\end{table}

\vspace{-2mm}
\paragraph{Utilizing Video Feature.}
To ensure a fair comparison, we utilize ResNet-50 as the feature extractor, in accordance with previous offline-GEBD studies.
However, since the baseline models were originally developed with different feature extractors and may underperform with ResNet-50 features. Table~\ref{tab:vidfeat} presents the results using the TSN network as a backbone, which was originally used by these baselines.
Despite utilizing different backbone architectures, our framework still outperforms all other baselines, thereby proving its robust effectiveness.

\vspace{-2mm}
\paragraph{Error Handling in OBD.}
A key challenge in On-GEBD is balancing sensitivity to upcoming boundaries while preventing false alarms.
OBD addresses this challenge by storing past errors to adjust its sensitivity to boundaries.
However, the persistence of elevated error values from a previous event boundary within the queue may lead to interference, hampering the detection of subsequent event boundaries.
To investigate this potential issue, we conducted an ablation study in which error values identified as event boundaries were excluded from the OBD queue.
As shown in Table~\ref{tab:obd_outlier_ablation}, excluding outliers severely degraded performance, reducing the Avg F1 from 0.748 to 0.663 (11.4\% relative drop).
This finding denotes that including errors from event boundaries is an essential mechanism for robust event boundary detection in OBD while ensuring computational efficiency.
We believe that one factor contributing to the effectiveness of this mechanism is its cognitive plausibility, as human perception adopts adaptive criteria for event segmentation when exposed to rapid changes \cite{Gordon2019}.

\subsection{Qualitative Result}
In Figure~\ref{fig:quali_1}, we present qualitative comparisons with the TeSTra-BC baseline. 
In contrast to the baseline, which shows noisy predictions as in all of the examples in Figure~\ref{fig:quali_1}, \method{} detects boundaries more accurately that align closely with the ground truth. 
Specifically, in the first example (upper panel), there are 5 ground-truth boundaries; \method{} identifies spikes at 4 of these boundaries within the gray-shaded ground-truth area, while TeSTra-BC detects only 3 boundaries—one of which falls inside an event segment rather than on an actual boundary. 
In the second example, which contains 7 ground-truth regions, TeSTra-BC successfully detects only 2 boundaries, whereas our method predicts 8 boundaries, 7 of which are correct.

\section{Conclusion}
\label{sec:conclusion}

We introduce a challenging new task, On-GEBD, designed to bring GEBD closer to human perceptual processes. 
To address this task, we present \method{}, inspired by Event Segmentation Theory (EST) from cognitive science, which explains how humans perceive and segment events. 
Our Consistent Event Anticipator (CEA) is trained using two losses—EST loss and REST loss—to effectively predict future frames consistent with current events, thereby maximizing errors at event boundaries. 
Additionally, the Online Boundary Discriminator (OBD) employs dynamic criteria to distinguish boundaries based on error values produced by the CEA. 
Our work demonstrates superior performance compared to baselines adapted from other online video understanding tasks and shows almost comparable performance to some recent work in the offline-GEBD task.

\paragraph{Acknowledgments.}
This work was supported by the IITP grants (RS-2019-II191842, RS-2021-II212068, RS2022-II220926 (30\%), RS-2022-II220077, RS-2022-II220113, RS-2022-II220959, RS-2022-II220871, RS-2025-02263598 (10\%), RS-2021-II211343 (SNU AI), RS-2021-II212068 (AI Innovation Hub), RS-2025-25442338 (AI Star Fellowship-SNU)) funded by MSIT, and the GIST-MIT Research Collaboration grant funded by GIST (10\%), Korea.
{
    \small
    \bibliographystyle{ieeenat_fullname}
    \bibliography{main}
}

\clearpage
\appendix
\setcounter{page}{1}
\maketitlesupplementary
\section{Ablation Study on Both $\tau$ and $\delta$ in OBD}

In Table~\ref{tab:obd_ablation}, we conduct an in-depth analysis on the interaction between the threshold $\tau$ and the queue size $\Delta$ for the Online Boundary Discriminator (OBD).
This table highlights the effects of varying these two parameters on the average F1 (Avg. F1) score in Kinetics-GEBD dataset~\cite{gebd_original}.
We observe that for a given queue size $\Delta$, increasing the threshold $\tau$ initially leads to improvements in performance up to a certain point, after which further increases in $\tau$ lead to a decline in the Avg. F1 score. 
For instance, when the queue size is fixed at $\Delta = 18$, the peak performance is achieved at $\tau = 1.5$, with an Avg. F1 score of 0.743. 
Increasing the threshold means selecting more severe outliers compared to the past errors stored in the OBD. 
Thus, setting a criterion that is either too strict or not would naturally result in a decline in overall performance.
We have determied $\tau$ as 1.5 throughout the entire experiment, since it demonstrates satisfactory performance as shown in the Table~\ref{tab:obd_ablation}.

Additionally, we can observe that performance gets better with lower $\tau$ values when  $\Delta$ increases.
For example, at a queue size of $\Delta = 24$, the highest F1 score is 0.751, which occurs at the lowest examined threshold of $\tau = 1.0$. 
This trend suggests that larger queues are better with lower thresholds, potentially due to the greater amount of past errors available in OBD queue when determining event boundaries.
We choose a queue size of $\Delta = 21$ and a threshold of $\tau = 1.5$, where the model achieves its optimal performance with an Avg. F1 score of 0.748. 

\begin{table}[t]
\centering
\footnotesize
\newcolumntype{Y}{>{\centering\arraybackslash}X}
    \caption{\textbf{Ablation study of Online Boundary Discriminator.} 
    Each number denotes an Avg. F1 score for the range of $\tau$ and $\Delta$.
    Bold number denotes the best Avg. F1 among the given same size of queue.
    }
    \begin{tabularx}{0.9\linewidth}{YYYYYY}
    \toprule
    \multirow{2}{*}{$\tau$} & \multicolumn{5}{c}{Size of Queue $\Delta$} \\ 
    \cmidrule{2-6}
      &   12   &   15   &   18   &   21  &  24 \\
    \midrule
    1.0 & 0.705 & 0.705 & 0.741 & 0.748 & \textbf{0.751} \\
    1.5 & 0.701 & 0.701 & \bf 0.743 & \textbf{0.748} & 0.747 \\
    2.0 & 0.715 & 0.715 & 0.740 & 0.735 & 0.726 \\
    2.5 & \textbf{0.731} & \textbf{0.728} & 0.716 & 0.701 & 0.684 \\
    3.0 & 0.711 & 0.691 & 0.670 & 0.649 & 0.629  \\
    \bottomrule
    \hline
    \end{tabularx}
\label{tab:obd_ablation}
\end{table}

\section{Further Experiments on $K$ in REST Loss}

The Regional EST (REST) loss is a core component in training our Consistent Event Anticipator (CEA), designed to enhance the model's ability to detect subtle changes at event boundaries.
The parameter $K$ determines the size of the temporal region considered in the REST loss calculation, controlling the range of frames that influences the loss computation.
To better understand the impact of this parameter, we conducted additional experiments varying the size of $K$, with results presented in Table~\ref{tab:windowsize_ablation}. 
These experiments reveal a clear trend in model performance as $K$ changes.
The Avg. F1 score shows a consistent increase as $K$ grows from 3 to 11, indicating that larger temporal context benefits the model's ability to detect event boundaries.
This improvement can be attributed to the model's enhanced capacity to capture longer-range dependencies and more complex temporal patterns within the video sequences.

Interestingly, our experimental result shows that the model's performance peaks when $K$ is set to 11 or 13, with both values yielding an Avg. F1 of 0.756. 
However, we observe a decline in performance for $K$ values beyond 13, suggesting that excessively large temporal regions may introduce noise or irrelevant information into the loss calculation.
Despite the highest performance at $K=11$ and $13$, we opted to use $K=9$ for all experiments reported in the main manuscript.
This decision was primarily due to practical considerations, considering the trade-off between model performance and computational resources.
Larger $K$ values require more GPU VRAM during training, which can limit batch sizes or necessitates more powerful hardware.

\begin{table}[t]
\centering
\scriptsize
\caption{\textbf{Ablation study of $K$ in REST loss.} Adjusting the range of REST loss in training CEA.}
\label{tab:windowsize_ablation}
\small
    \begin{tabular}{@{\hspace{1.5pt}}c@{\hspace{1.5pt}}c@{\hspace{1.5pt}}c@{\hspace{1.5pt}}c@{\hspace{1.5pt}}c@{\hspace{1.5pt}}c@{\hspace{1.5pt}}c@{\hspace{1.5pt}}c@{\hspace{1.5pt}}c@{\hspace{1.5pt}}c}
    \toprule
    \scriptsize $K$ &   \scriptsize 3   &   \scriptsize 5   &   \scriptsize 7   &   \scriptsize 9  & \scriptsize 11 & \scriptsize 13 & \scriptsize 15 & \scriptsize 17& \scriptsize 19\\
    \midrule
    \scriptsize{Avg F1}      &~\scriptsize{0.724}~  &~\scriptsize{0.733}~  &~\scriptsize{0.743}~ &~\scriptsize{0.748}~ & \bf ~\scriptsize{0.756}~ & \bf ~\scriptsize{0.756}~ & ~\scriptsize{0.754}~ & ~\scriptsize{0.749}~& ~\scriptsize{0.746}~ \\
    \bottomrule
    \end{tabular}
\end{table}
\begin{table}[t]
    \centering
    \footnotesize
    \setlength{\tabcolsep}{20pt}

    \caption{\textbf{Comparison of different lengths, Avg F1 scores, and VRAM usage.} We denote the highest Avg F1 in \textbf{bold}.}
    \resizebox{0.95\columnwidth}{!}{
    \begin{tabular}{ccc}
        \toprule
        \multirow{1}{*}{Length} & \multirow{1}{*}{Avg F1} & \multirow{1}{*}{VRAM (GB)} \\
        \cmidrule(lr){1-1} \cmidrule(lr){2-2} \cmidrule(lr){3-3}
        4 & 0.728 & 5.2\\
        8 & \textbf{(Ours) 0.748} & \textbf{9.0} \\
        16 & 0.742 & 14.8 \\
        32 & 0.745 & 27.9  \\
        \bottomrule
     \end{tabular}
    }
    \vspace{-1.5em}
    \label{tab:length_comparison}
\end{table}

\begin{table*}[t]
    \centering
    \caption{\textbf{Quantitative comparison with additional offline methods.} In addition to the offline GEBD methods presented in Table~2 of our original manuscript, we include additional results from more recent offline approaches to highlight the robustness of our model, even as an online method. Note that we report the performance of the models in an offline setting from their original literature.
    Also, we indicate the highest F1 score with \textbf{bold} for each dataset. 
    }
    \resizebox{\textwidth}{!}{
    \begin{tabular}{cccccccccccccccc}
    \hline
    \toprule
    Dataset&Setting&Supervision&Rel. Dis. threshold & 0.05 & 0.1 & 0.15 & 0.2 & 0.25 & 0.3 & 0.35 & 0.4 & 0.45 & 0.5 & Avg \\
    \midrule
    \multirow{16}{*}{\rotatebox[origin=c]{90}{Kinetics-GEBD}}&\multirow{15}{*}{Offline}&\multirow{10}{*}{Supervised}& BMN~\cite{bmn} & 0.186 & 0.204 & 0.213 & 0.220 & 0.226 & 0.230 & 0.233 & 0.237 & 0.239 & 0.241& 0.223\\
    &&& BMN-StartEnd~\cite{gebd_original}&0.491 & 0.589 & 0.627 & 0.648 & 0.660 & 0.668 & 0.674 & 0.678 & 0.681 & 0.683& 0.640\\
    &&& TCN-TAPOS~\cite{gebd_original}& 0.464 & 0.560 & 0.602 & 0.628 & 0.645 & 0.659 & 0.669 & 0.676 & 0.682 & 0.687 & 0.627\\
    &&& TCN~\cite{TCN}&0.588 & 0.657 & 0.679 & 0.691 & 0.698 & 0.703 & 0.706 & 0.708 & 0.710 & 0.712 & 0.685\\
    &&& PC~\cite{gebd_original} & 0.625 & 0.758 & 0.804 & 0.829 & 0.844 & 0.853 & 0.859 & 0.864 & 0.867 & 0.870 & 0.817\\
    
    &&& Temporal Perceiver~\cite{temporal_perceiver} & 0.748 & 0.828 & 0.852 & 0.866 & 0.874 & 0.879 & 0.883 & 0.887 & 0.890 & 0.892 & 0.860 \\
    &&& SBoCo-Res50~\cite{uboco} &  0.732& -& -& -& -& -& -& -& -& -& 0.866\\
    &&& DDM-Net~\cite{ddm_net}& 0.764 & 0.843 & 0.866 & 0.880 & 0.887 & 0.892 & 0.895 & 0.898 & 0.900 & 0.902& 0.873 \\
    &&& SC-Transformer~\cite{sc-transformer} & 0.777 & 0.849 & 0.873 & 0.886 & 0.895 & 0.900 & 0.904 & 0.907 & 0.909 & 0.911 & 0.881 \\
    &&& EfficientGEBD~\cite{efficientgebd} &  0.783 &  0.851 & - & - & - & 0.901 & - & - & - & 0.913 & 0.883 \\
    &&& LCVSL~\cite{lcvsl} &  0.768 &  0.848 & 0.872 & 0.885 & 0.892 & 0.896 & 0.899 & 0.901 & 0.903 & 0.906 & 0.877 \\
    
    &&& DyBDet~\cite{dybnet} & \bf 0.796 & \bf 0.858 & \bf 0.880 & \bf 0.893 & \bf 0.901 & \bf 0.907 & \bf 0.911 & \bf 0.915 & \bf 0.917 & \bf 0.919 & \bf 0.890 \\
    \cmidrule{3-15}
    &&\multirow{5}{*}{Unsupervised} & SceneDetect~\cite{gebd_original} &0.275 & 0.300 & 0.312 & 0.319 & 0.324 & 0.327 & 0.330 & 0.332 & 0.334 & 0.335 & 0.318 \\
    &&& PA-Random~\cite{gebd_original} &0.336 & 0.435 & 0.484 & 0.512 & 0.529 & 0.541 & 0.548 & 0.554 & 0.558 & 0.561& 0.506 \\
    &&& PA~\cite{gebd_original} &0.396 & 0.488 & 0.520 & 0.534 & 0.544 & 0.550 & 0.555 & 0.558 & 0.561 & 0.564 & 0.527\\
    &&& CoSeg~\cite{coseg} & 0.656 & 0.758 & 0.783 & 0.794 & 0.799 & 0.803 &0.804 & 0.806 & 0.807 & 0.809 & 0.782 \\
    &&& UBoCo-Res50~\cite{uboco} & 0.703& -& -& -& -& -& -& -& -& -& 0.867\\
    &&& FlowGEBD~\cite{flowgebd} & 0.713& 0.828& 0.850& 0.858& 0.862& 0.864& 0.866& 0.867& 0.868& 0.869& 0.845\\
    
\cmidrule{2-15}
   &Online&Supervised& \textbf{\method{} (Ours)} &  0.620 &  0.687 & 0.724 &  0.746 & 0.762 &  0.774 &  0.782 &  0.789 &   0.795 &   0.799 &  \ 0.748 \\
\midrule
\midrule
       \multirow{12.5}{*}{\rotatebox[origin=c]{90}{TAPOS}}&\multirow{11}{*}{Offline}&\multirow{9}{*}{Supervised}&ISBA~\cite{gebd_original} & 0.106 & 0.170 & 0.227 & 0.265 & 0.298 & 0.326 & 0.348 & 0.348 & 0.348 & 0.348 & 0.330\\
    &&& TCN~\cite{gebd_original} & 0.237 & 0.312 & 0.331 & 0.339 & 0.342 & 0.344 & 0.347 & 0.348 & 0.348 & 0.348& 0.330\\
    &&& CTM~\cite{gebd_original}& 0.244 & 0.312 & 0.336 & 0.351 & 0.361 & 0.369 & 0.374 & 0.381 & 0.383 & 0.385 & 0.350\\
    &&& TransParser~\cite{TCN} & 0.289 & 0.381 & 0.435 & 0.475 & 0.500 & 0.514 & 0.527 & 0.534 & 0.540 & 0.545 & 0.474\\
    &&& PC~\cite{gebd_original} & 0.522 & 0.595 & 0.628 & 0.646 & 0.659 & 0.665 & 0.671 & 0.676 & 0.679 & 0.683 & 0.642\\
    &&& DDM-Net~\cite{ddm_net}& 0.604 & 0.681 & 0.715 & 0.735 & 0.747 & 0.753 & 0.757 & 0.760 & 0.763 & 0.767 & 0.728 \\
    &&& Temporal Perceiver~\cite{temporal_perceiver} & 0.552 & 0.663 & 0.713 & 0.738 & 0.757 & 0.765 & 0.774 & 0.779 & 0.784 & 0.788 & 0.732 \\
    &&& SC-Transformer~\cite{sc-transformer} & 0.618 & 0.694 & 0.728 & 0.749 & 0.761 & 0.767 & 0.771 & 0.774 & 0.777 & 0.780 & 0.742 \\
    &&& EfficientGEBD~\cite{efficientgebd} &  0.631 &  0.705 & - & - & - & 0.774 & - & - & - & 0.786 & 0.748 \\
    &&& LCVSL~\cite{lcvsl} &  0.618 &  0.694 & 0.728 & 0.749 & 0.761 & 0.767 & 0.771 & 0.774 & 0.777 & 0.780 & 0.742 \\
    &&& DyBDet~\cite{dybnet} & \bf 0.625 & \bf 0.701 & \bf 0.734 & \bf 0.756 & \bf 0.767 & \bf 0.772 & \bf 0.775 & \bf 0.779 & \bf 0.781 & \bf 0.784 & \bf 0.747 \\
    \cmidrule{3-15}
    &&\multirow{3}{*}{Unsupervised} & SceneDetect~\cite{gebd_original} &0.035 & 0.045 & 0.047 & 0.051 & 0.053 & 0.054 & 0.055 & 0.056 & 0.057 & 0.058 & 0.051 \\
    &&& PA-Random~\cite{gebd_original} &0.158 & 0.233 & 0.273 & 0.310 & 0.331 & 0.347 & 0.357 & 0.369 & 0.376 & 0.384 & 0.314 \\
    &&& PA~\cite{gebd_original} & 0.360 & 0.459 & 0.507 & 0.543 & 0.567 & 0.579 &  0.592  &  0.601  &  0.609  &  0.615  & 0.543 \\
    &&& FlowGEBD~\cite{flowgebd} & 0.375& 0.502& 0.569& 0.624& 0.658& 0.677& 0.695& 0.703& 0.711& 0.717& 0.623\\

    \cmidrule{2-15}
    &Online&Supervised& \textbf{\method{} (Ours)} &  0.394  & 0.455 & 0.499 & 0.532  &  0.558  &  0.578 &  0.594 & 0.608 &0.619 &0.629 & 0.547 \\
    \bottomrule
    \hline
    \end{tabular}
    }
    \label{tab:supp_offline_online_comp}
\end{table*}

\section{Ablation on Length $L$}
The choice of input video sequence length impacts both the performance and computational efficiency of our model. 
A longer input sequence provides more temporal context, potentially improving boundary detection accuracy but at the cost of increased VRAM consumption and inference time. 
Conversely, shorter sequences are computationally efficient but may lack sufficient context for detecting subtle event transitions.

To achieve a balance between performance and efficiency, we set the input length to an optimal value based on empirical results. As shown in Table~\ref{tab:length_comparison}, we compare different sequence lengths in terms of Avg F1 score and VRAM usage. 
Our selected input length achieves the highest Avg F1 score while maintaining a reasonable VRAM footprint, making it suitable for real-time processing.

Our OBD is designed to dynamically adapt to recent boundary patterns, reducing false positives during frequent changes while maintaining sensitivity in stable periods. 
This design aligns with human perception, as studies suggest that when individuals are exposed to rapidly changing visuals, they naturally adjust their threshold for identifying meaningful event boundaries \cite{Gordon2019}. 
The ability to incorporate past outliers ensures that the model remains adaptable to varying event structures without excessive desensitization to new transitions.

These findings reinforce the necessity of including outliers in the queue to maintain robust event boundary detection, making our approach both computationally effective and cognitively plausible.

\begin{table}[t]
     \centering
        \caption{\textbf{Ablation on batch-wise weighted loss.}}
        \resizebox{0.45\columnwidth}{!}{
        \begin{tabular}{cc}
        \toprule
        Batch-wise loss  &  ~~Avg F1~~  \\
        \midrule
        \ding{55} & 0.743 \\
        \ding{51} & \textbf{0.748}\\
        \bottomrule
        \end{tabular}}
        \label{tab:batchloss} 
\end{table}

\begin{table*}[t!]
  \centering
  \caption{\textbf{Quantitative comparison for generalization ability.} Results on Youtube-INRIA-Instructional dataset with online and offline baselines.}
  \resizebox{0.75\textwidth}{!}{%
    \begin{tabular}{cccccc}
      \toprule
      \multirow{1}{*}{~~~~Online~~~} & \multirow{1}{*}{~~~~~Method~~~} & \multirow{1}{*}{~~Pretrained~~} & \multirow{1}{*}{~~Precision@0.05~~} & \multirow{1}{*}{~~Recall@0.05~~} & \multirow{1}{*}{~~F1@0.05~~}\\
      \cmidrule(lr){1-1} \cmidrule(lr){2-2} \cmidrule(lr){3-3} \cmidrule(lr){4-4} \cmidrule(lr){5-5} \cmidrule(lr){6-6}
      \multirow{2}{*}{X} & U-Net\footnotemark[3]   & INRIA &   -   &   -   & 0.299 \\
      & CoSeg [41] & INRIA & \textbf{0.467} & \textbf{0.633} & \textbf{0.537} \\
      \midrule
       \multirow{5.5}{*}{O} & TeSTra – BC   & Kinetics-GEBD &  0.181 & 0.748 & 0.291\\
        & Sim-On – BC   & Kinetics-GEBD & 0.099	& 0.068 & 0.080 \\
        & OadTR – BC    & Kinetics-GEBD & \underline{0.348} & \underline{0.526} & \underline{0.419}\\
        & MiniROAD - BC & Kinetics-GEBD & 0.209 & 0.572 & 0.306 \\
        \cmidrule{2-6}
            & Ours          & Kinetics-GEBD & \textbf{0.411} & \textbf{0.666} & \textbf{0.508} \\
      \bottomrule
    \end{tabular}
  }
  \label{tab:inria_dataset}
  \vspace{-1.3em}
\end{table*}

\section{Additional Offline GEBD performance table}
We further report the performance of models developed and evaluated under an offline setting in Table~\ref{tab:supp_offline_online_comp}.
Compared to the Table 2 in our main manuscript, Table~\ref{tab:supp_offline_online_comp} additionally include Temporal Perceiver~\cite{temporal_perceiver}, SBoCo-Res50~\cite{uboco}, DDM-Net~\cite{ddm_net}, SC-Transformer~\cite{sc-transformer}, UBoCo~\cite{uboco}, Efficient-GEBD~\cite{efficientgebd}, LCVSL~\cite{lcvsl}, DyBDet~\cite{dybnet} and FlowGEBD~\cite{flowgebd} for the Kinetics-GEBD dataset. 
For the TAPOS dataset, we have additionally included DDM-Net, Temporal Perceiver, SC-Transformer, Efficient-GEBD~\cite{efficientgebd}, LCVSL~\cite{lcvsl}, DyBDet and FlowGEBD~\cite{flowgebd}  as UBoCo do not report performance for this dataset.

\section{Ablation on Batch-wise Weighted Loss}

Table~\ref{tab:batchloss} presents the Avg. F1 score on the Kinetics-GEBD dataset, evaluating the impact of batch-wise weighted loss in our model.
This technique addresses the imbalance between boundary and non-boundary frames in the training data, a common challenge in event boundary detection tasks.
By dynamically adjusting the importance of samples within a single batch during training, the batch-wise weighted loss aims to improve the model's sensitivity to boundary frames without manual hyper-parameter tuning.

The results indicate that incorporating batch-wise weighted loss yields a 0.5\%p increase in the Avg. F1 score.
This improvement may seem trivial, but considering the sensitivity of detecting generic event boundaries, we conjecture that batch-wise weighting is showing noticeable improvement in accuracy.

\section{Zero-shot Ability of Our Framework}
To further demonstrate the generalization capability of our framework, we evaluate our framework on the challenging YouTube-INRIA-Instructional dataset~\cite{inria} (Table~\ref{tab:inria_dataset}), which was used in~\cite{coseg} and consists of long-form, multi-minute instructional videos—markedly different in nature from Kinetics-GEBD.
Without any additional finetuning, our model pretrained solely on Kinetics-GEBD achieves an F1@0.05 score of 0.508.
This result is competitive with, or even superior to, existing offline methods, and it consistently outperforms all online baselines.
These results highlight the strong zero-shot generalization ability of our model to previously unseen, complex video domains.

\section{Additional Details on Computational Cost}
In Table~\ref{tab:fps} of the main manuscript, we analyze the real-time performance of our proposing model, focusing on its inference speed (\ie FPS).
For completeness, we provide additional real-time metrics including computational cost details (\eg, GFLOPs and memory usage) in Table~\ref{tab:comp_cost}, highlighting the efficiency of our method in online scenario.
As showcased in the Table~\ref{tab:comp_cost}, our model achieves best performance despite having compatible number of GFLOPs and parameters compared to the most efficient baselines (\ie, Sim-On-BC, MiniROAD-BC), demonstrating the effectiveness.

\begin{table*}[t!]
  \caption{\textbf{Comparison of real-time performance with computational cost.} Note that \textbf{bold} refers to the best and \underline{underline} refers to the second best.}
  \centering
  \resizebox{0.75\textwidth}{!}{%
    \begin{tabular}{lcccccc}
      \toprule
      \multirow{1}{*}{Method} & \multirow{1}{*}{~~\# of param.~~} & \multirow{1}{*}{~~GFLOPs~~$\downarrow$} & \multirow{1}{*}{~VRAM (MB)~$\downarrow$} & \multirow{1}{*}{~FPS~$\uparrow$} & \multirow{1}{*}{~Avg. F1~~$\uparrow$}\\
      \cmidrule(lr){1-1} \cmidrule(lr){2-2} \cmidrule(lr){3-3} \cmidrule(lr){4-4} \cmidrule(lr){5-5}  \cmidrule(lr){6-6}
      TeSTra – BC   & 48.73M & 17.0 & 354 & 72.5 & 0.557\\
      Sim-On – BC   & 24.70M & \textbf{8.2} & \textbf{134} & 76.3 & 0.618\\
      OadTR – BC    & 97.10M & 13.0 & 385 & 48.9 & 0.558\\
      MiniROAD - BC & 37.15M & \textbf{8.2} & \textbf{134} & \textbf{99.8} & \underline{0.681}\\
      \midrule
      Ours    & 42.41M & \underline{10.3} & \underline{228} & \underline{96.3} & \textbf{0.748}\\
      \bottomrule
    \end{tabular}
  }
        \vspace{0.15 em}
    {\\ \scriptsize $*$All experiments were conducted on a single NVIDIA RTX~A6000 GPU.}
  \label{tab:comp_cost}
\end{table*}

\section{Additional Qualitative Result}
\begin{figure*}[t!]
    \centering
    \includegraphics[width=0.98\textwidth]{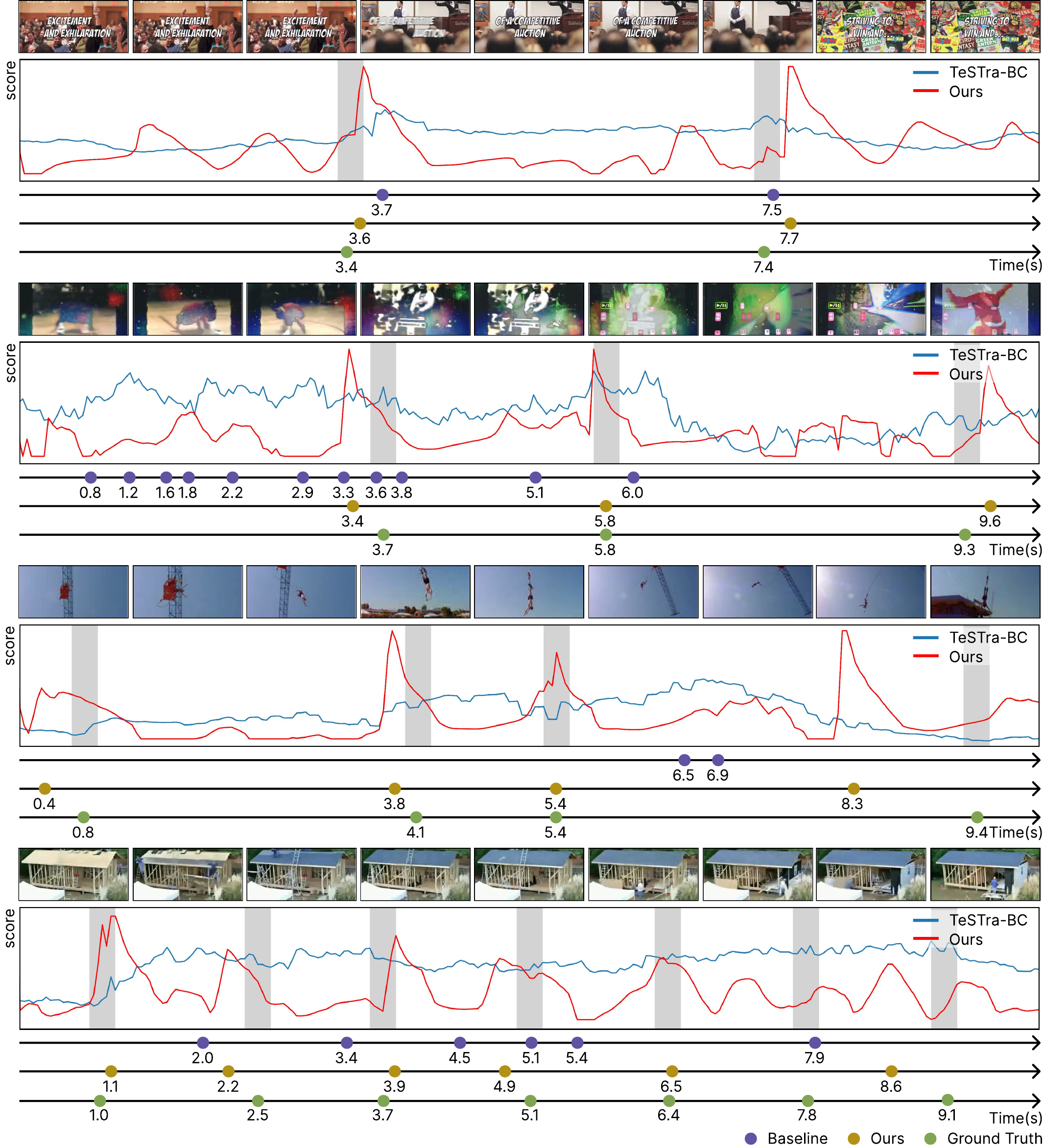} 
        \caption{\textbf{Additional qualitative result on Kinetics-GEBD dataset.} Comparison between our proposed framework and the baseline (TeSTra-BC~\cite{testra}).
    }
    \label{fig:add_quali_k400}
\end{figure*}

We illustrate more qualitative results of our model compared to one of baselines (TeSTra-BC~\cite{testra}), on both Kinetics-GEBD and TAPOS~\cite{tapos} datasets.
In Figure~\ref{fig:add_quali_k400}, we present two cases of abrupt scene changes (\ie, first and second row) and two cases of subtle changes (\ie, third and fourth row) in Kinetics-GEBD dataset.

The first row shows a distinct transition such as shot changes between events in a video. 
In this straightforward scenario, both the baseline and our method yield results that are close to the ground truth. 
However, the error plot of our method for each frame shows sharp peaks, distinctively indicating the boundary locations, in contrast to the baseline's, which presents a nearly flat distribution.
In the second row, there are changes of scene not only at event boundaries but also within each event.
While TeSTra-BC fails to recognize the semantic continuity at the first event of the video and raises numerous false alarms, our framework recognizes the boundaries successfully.
The third and fourth example present cases where the transition of events is subtle, requiring a deeper understanding of granular details to detect event boundaries.
Our model also outperforms the baseline in identifying event boundaries.

\begin{figure*}[t!]
    \centering
    \includegraphics[width=0.98\textwidth]{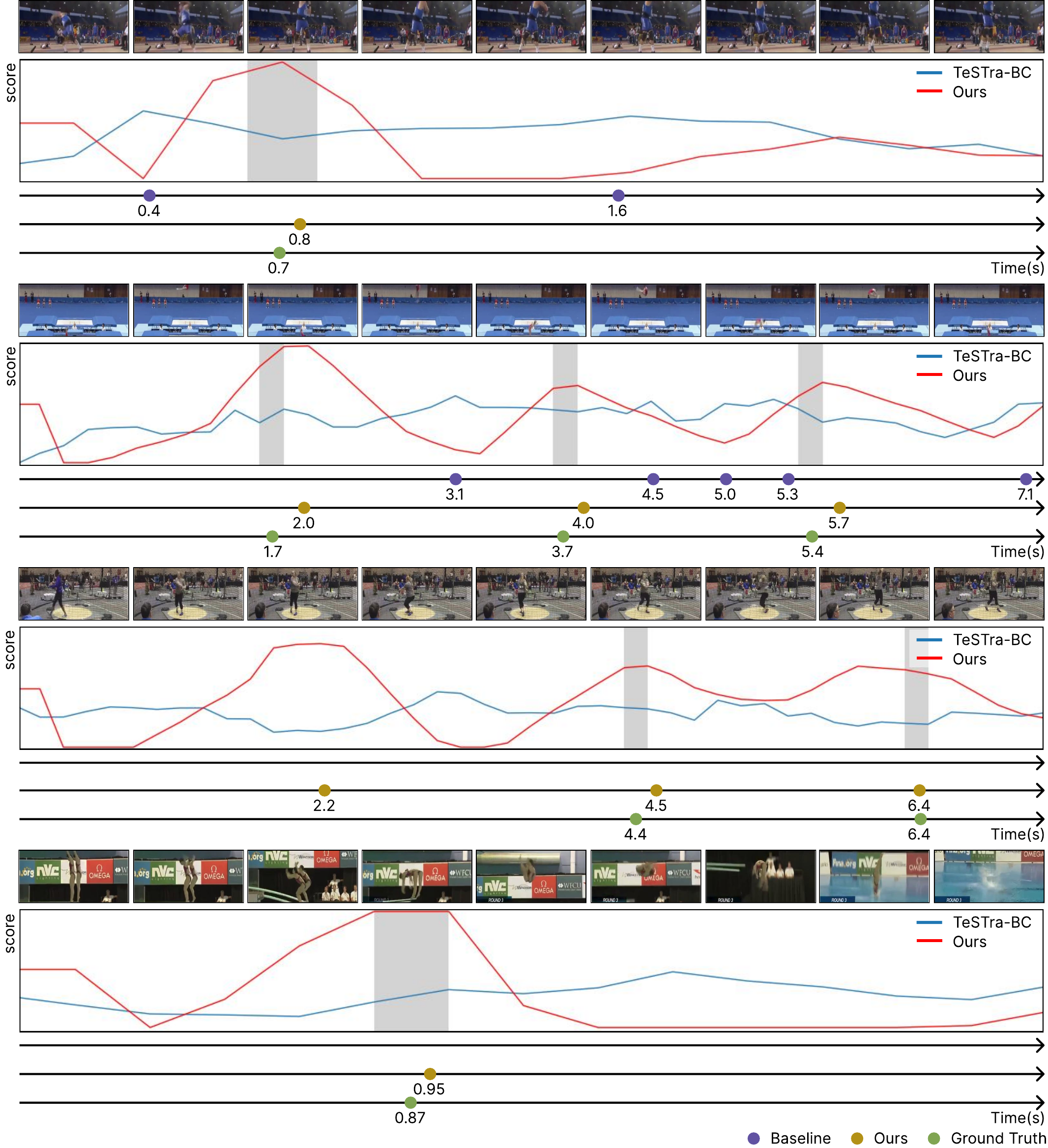} 
        \caption{\textbf{Additional qualitative result on TAPOS dataset.} Comparison between our proposed framework and the baseline (TeSTra-BC~\cite{testra}).
    }
    \label{fig:add_quali_tapos}
\end{figure*}

In Figure~\ref{fig:add_quali_tapos}, we present a comparison between TeSTra-BC and our framework on the TAPOS dataset.
As mentioned in our main manuscript, the TAPOS dataset consists of Olympic sport videos annotated with 21 action classes, where each action is further divided into multiple sub-actions.
Since these sub-actions are re-purposed as a single event in our experiment, the semantic changes between sub-actions within the single video tend to be subtle.
As shown in Figure~\ref{fig:add_quali_tapos}, TeSTra-BC fails to detect event boundaries in all four cases, particularly failing to detect any boundaries in the third and fourth cases.
In contrast, our framework successfully detects the subtle semantic changes occurring at event boundaries in all videos.

\section{Limitation and Social Impact}
Although the Kinetics-GEBD and TAPOS dataset are the only datasets available for testing the GEBD task, they consist exclusively of sports or exercise-related videos.
In this context, OBD, which introduces a novel criterion for defining event boundaries, may exhibit bias toward sports or exercise contexts.
To ensure robust performance across a diverse range of domains, it may be necessary to construct a variety of datasets for GEBD and perform a tuning of corresponding parameters (\eg, $\Delta$, $\tau$).

Since the On-GEBD solver is able to process diverse long-form videos in real time, it has the potential to impact fields that require continuous monitoring and rapid analysis within the previously unobserved video streams such as public safety and surveillance.

\end{document}